\documentclass[preprint,12pt,authoryear]{elsarticle}

\usepackage{amsmath,graphicx}
\usepackage{pgfplots}
\usepackage{amssymb}
\usepackage{amsthm}
\usepackage{algorithm}
\usepackage{algorithmic}
\usepackage{subfigure}
\usepackage[pagebackref=true,breaklinks=true,letterpaper=true,colorlinks,bookmarks=false]{hyperref}
\journal{Neurocomputing}

\begin{document}

\begin{frontmatter}

%% Title, authors and addresses

%% use the tnoteref command within \title for footnotes;
%% use the tnotetext command for theassociated footnote;
%% use the fnref command within \author or \address for footnotes;
%% use the fntext command for theassociated footnote;
%% use the corref command within \author for corresponding author footnotes;
%% use the cortext command for theassociated footnote;
%% use the ead command for the email address,
%% and the form \ead[url] for the home page:
%% \title{Title\tnoteref{label1}}
%% \tnotetext[label1]{}
%% \author{Name\corref{cor1}\fnref{label2}}
%% \ead{email address}
%% \ead[url]{home page}
%% \fntext[label2]{}
%% \cortext[cor1]{}
%% \address{Address\fnref{label3}}
%% \fntext[label3]{}

\title{Label Embedded Dictionary Learning for Image Classification}

%% use optional labels to link authors explicitly to addresses:
%% \author[label1,label2]{}
%% \address[label1]{}
%% \address[label2]{}

\author[UPC]{Shuai Shao}
\ead{shuaishao@s.upc.edu.cn}
\author[UPC]{Yan-Jiang Wang\corref{cor1}}
\cortext[cor1]{PaperID:NEUCOM-S-19-01925 Corresponding author.}
\ead{yjwang@upc.edu.cn}
\author[UPC]{Bao-Di Liu\corref{cor2}}
\cortext[cor2]{PaperID:NEUCOM-S-19-01925 Corresponding author.}
\ead{thu.liubaodi@gmail.com}
\author[UPC]{Weifeng Liu}
\ead{liuwf@upc.edu.cn}
\author[UPC]{Rui Xu}
\ead{xddxxr@126.com}

\address[UPC]{College of Information and Control Engineering, China University of Petroleum, Qingdao 266580, China}

\begin{abstract}
%% Text of abstract
Recently, label consistent k-svd (LC-KSVD) algorithm has been successfully applied in image classification. 
The objective function of LC-KSVD is consisted of reconstruction error, classification error and discriminative sparse codes error with $\ell_0$-norm sparse regularization term. 
The $\ell_0$-norm, however, leads to NP-hard problem. 
Despite some methods such as orthogonal matching pursuit can help solve this problem to some extent, it is quite difficult to find the optimum sparse solution. 
To overcome this limitation, we propose a label embedded dictionary learning (LEDL) method to utilise the $\ell_1$-norm as the sparse regularization term so that we can avoid the hard-to-optimize problem by solving the convex optimization problem.
Alternating direction method of multipliers and blockwise coordinate descent algorithm are then exploited to optimize the corresponding objective function. 
Extensive experimental results on six benchmark datasets illustrate that the proposed algorithm has achieved superior performance compared to some conventional classification algorithms.
\end{abstract}

\begin{keyword}
Dictionary learning \sep sparse representation \sep label embedded dictionary learning \sep image classification
%% keywords here, in the form: keyword \sep keyword

%% PACS codes here, in the form: \PACS code \sep code

%% MSC codes here, in the form: \MSC code \sep code
%% or \MSC[2008] code \sep code (2000 is the default)

\end{keyword}

\end{frontmatter}

%% \linenumbers

%% main text
\section{Introduction}
\label{Introduction}

Recent years, image classification has been a classical issue in computer vision. Many successful algorithms~\cite{yu2012adaptive,shi2018hypergraph,wright2009robust,yu2014high,song2018euler,yu2013pairwise,wang2018iterative,yu2012image,yang2009linear,yang2017discriminative,liu2014class,liu2017class,jiang2013label,hao2017class,chan2015pcanet,nakazawa2018wafer,ji2014spectral,xu2019sparse,yuan2016non} have been proposed to solve the problem. In these algorithms, there is one category that contributes a lot for image classification which is the sparse representation based method.

\begin{figure*}
	\begin{center}
		\includegraphics[width=1.0\linewidth]{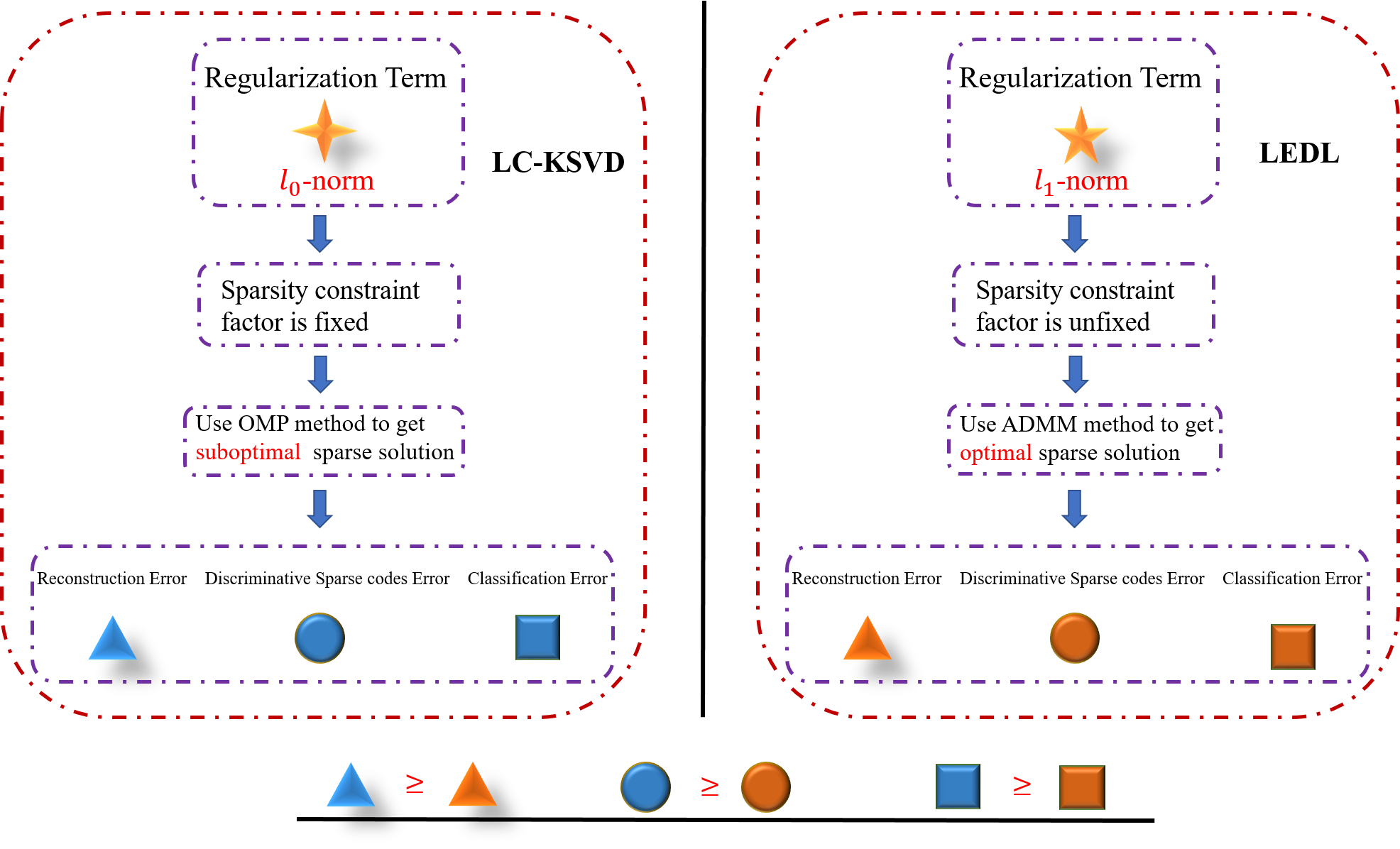}
	\end{center}
	\caption{
		The scheme of LEDL is on the right while the LC-KSVD is on the left. The difference between the two methods is the sparse regularization term which LEDL use the $\ell_1$-norm regularization term and LC-KSVD use the $\ell_0$-norm regularization term. Compared with $\ell_0$-norm, the sparsity constraint factor of $\ell_1$-norm is unfixed so that the basis vectors can be selected freely for linear fitting. Thus, our proposed LEDL method can get smaller errors than LC-KSVD. 
	}
	\label{fig:Comparision}
\end{figure*}

Sparse representation is capable of expressing the input sample features as a linear combination of atoms in an overcomplete basis set. \cite{wright2009robust} proposed sparse representation based classification (SRC) algorithm which use the $\ell_1$-norm regularization term to achieve impressive performance. 
SRC is the most representative one in the sparse representation based methods.
However, in traditional sparse representation based methods, training sample features are directly exploited without considering the discriminative information which is crucial in real applications.
That is to say, sparse representation based methods can gain better performance if the discriminative information is properly harnessed. 

To handle this problem, dictionary learning (DL) method is introduced to preprocess the training sample features befor classification. 
DL is a generative model for sparse representation which the concept was firstly prposed by~\cite{mallat1993matching}.
A few years later, \cite{olshausen1996emergence,olshausen1997sparse} proposed the application of DL on natural images and then it has been widely used in many fields such as image  denoising~\cite{chang2000adaptive,li2012efficient,li2018joint}, image superresolution~\cite{yang2010image,wang2012semi,gao2018self}, and image classification~\cite{liu2017class,jiang2013label,chang2016learning}.
A well learned dictionary can help to get significant boost in classification accuracy. 
Therefore, DL based methods in classification are more and more popular in recent years.

Specificially, there are two strategies are proposed to successfully utilise the discriminative information: i) class specific dictionary learning ii) class shared dictionary learning. The first strategy is to learn specific dictionaries for each class such as~\cite{wang2012supervised,yang2014sparse,liu2016face}. The second strategy is to learn a shared dictionary for all classes. For example,~\cite{zhang2010discriminative} proposed discriminative K-SVD(D-KSVD) algorithm to directly add the discriminative information into objective function. Furthermore,~\cite{jiang2013label} proposed label consistence K-SVD (LC-KSVD) method which add a label consistence term into the objective function of D-KSVD. The motivation for adding this term is to encourage the training samples from the same class to have similar sparse codes and those from different classes to have dissimilar sparse codes. Thus, the discriminative abilities of the learned dictionary is effectively improved. However, the sparse regularization term in LC-KSVD is $\ell_0$-norm which leads to the NP-hard~\cite{natarajan1995sparse} problem. Although some greedy methods such as orthogonal matching pursuit (OMP)~\cite{tropp2007signal} can help solve this problem to some extent, it is usually to find the suboptimum sparse solution instead of the optimal sparse solution. 
More specifically, greedy method solve the global optimal problems by finding basis vectors in order of reconstruction errors from small to large until $T$ (the sparsity constraint factor) times. Thus, the initialized values are crucial. To this end, $\ell_0$-norm based sparse constraint is not conducive to finding a global minimum value to obtain the optimal sparse solution.

In this paper, we propose a novel dictionary learning algorithm named label embedded dictionary learning (LEDL). This method introduces the $\ell_1$-norm regularization term to replace the $\ell_0$-norm regularization of LC-KSVD. Thus, we can freely select the basis vectors for linear fitting to get optimal sparse solution. In addition, $\ell_1$-norm sparse representation is widely used in many fields so that our proposed LEDL method can be extended and applied easily. We show the difference between our proposed LEDL and LC-KSVD in Figure~\ref{fig:Comparision}. We adopt the alternating direction method of multipliers (ADMM)~\cite{boyd2011distributed} framework and blockwise coordinate descent (BCD)~\cite{liu2014blockwise} algorithm to optimize LEDL. Our work mainly focuses on threefold.

\begin{itemize}	
	\item We propose a novel dictionary learning algorithm named label embedded dictionary learning which introduces the $\ell_1$-norm regularization term as the sparse constraint. The $\ell_1$-norm sparse constraint is able to help easily find the optimal sparse solution.
	
	\item We propose to utilize the alternating direction method of multipliers (ADMM)~\cite{boyd2011distributed} algorithm and blockwise coordinate descent (BCD)~\cite{liu2014blockwise} algorithm to optimize dictionary learning task.
	
	\item We verify the superior performance of our method on six benchmark datasets.

\end{itemize}

The rest of the paper is organized as follows. Section ~\ref{Related work} reviews two conventional methods which are SRC and LC-KSVD. 
Section~\ref{Methodology-A} presents LEDL method for image classification. The optimization approach and the convergence are elaborated in Section~\ref{Methodology-B}. Section~\ref{Experimental results} shows experimental results on six well-known datasets. Finally, we conclude this paper in Section~\ref{Conclusion}.

\section{Related Work}\label{Related work}

In this section, we overview two related algorithms, including sparse representation based classification (SRC) and label consistent K-SVD (LC-KSVD).

\subsection{Sparse representation based classification (SRC)}

SRC was proposed by~\cite{wright2009robust}. Assume that we have $C$ classes of training samples, denoted by $ {{\mathbf{X}}_{c}},c=1,2,\cdots ,C$, where ${{\mathbf{X}}_{c}}$ is the training sample matrix of class $c$. 
Each column of the matrix ${{\mathbf{X}}_{c}}$ is a training sample feature from the $c_{th}$ class. 
The whole training sample matrix can be denoted as $\mathbf{X}=\left[ {{\mathbf{X}}_{1}},{{\mathbf{X}}_{2}},\cdots {{\mathbf{X}}_{C}} \right]\in {{\mathbb{R}}^{D\times N}}$, where $D$ represents the dimensions of the sample features and $N$ is the number of training samples. Supposing that $\mathbf{y}\in {{\mathbb{R}}^{D\times 1}}$ is a testing sample vector, the sparse representation algorithm aims to solve the following objective function:
\begin{equation}
	\begin{split}
		{\bf{\hat s}} = \arg {\min _{\bf{s}}}{\mkern 1mu} \left\{ {\left\| {{\bf{y}} - {\bf{Xs}}} \right\|_2^2 + 2\alpha {{\left\| {\bf{s}} \right\|}_1}} \right\}
	\end{split}\label{SRC}
\end{equation}
where, $\alpha$ is the regularization parameter to control the tradeoff between fitting goodness and sparseness. The sparse representation based classification is to find the minimum value of the residual error for each class.
\begin{equation}
	\begin{split}
		id\left( {\bf{y}} \right) = \arg {\min _c}{\mkern 1mu} \left\| {{\bf{y}} - {{\bf{x}}_c}{{{\bf{\hat s}}}_c}} \right\|_2^2
	\end{split}\label{residual}
\end{equation}
where $id\left( \bf{y} \right)$ represents the predictive label of $\bf{y}$, ${{{\bf{\hat s}}}_c}$ is the sparse code of $c_{th}$ class. The procedure of SRC is shown in Algorithm~\ref{Algorithm1}. 
Obviously, the residual $e_c$ is associated with only a few images in class $c$. 

\begin{algorithm}[!t]
	\scriptsize
	\caption{Sparse representation based classification}\label{Algorithm1}
	\hspace*{0.02in} {\bf Input:} ${\bf{X}}\in\mathbb{R}^{D\times N}$, ${\bf{y}}\in\mathbb{R}^{D\times 1}$, $\alpha$\\
	\hspace*{0.02in} {\bf Output:} $id({\bf{y}})$
	\begin{algorithmic}[1]
		\STATE Code ${\bf{y}}$ with the dictionary ${\bf{X}}$ via $\ell_1$-minimization. \STATE ${\bf{\hat s}} = \arg {\min _{\bf{s}}}{\mkern 1mu} \left\{ {\left\| {{\bf{y}} - {\bf{Xs}}} \right\|_2^2 + 2\alpha {{\left\| {\bf{s}} \right\|}_1}} \right\}$
		\FOR{$c=1$;$c\le C$;$c\!+\!+$}
		\STATE Compute the residual ${e_c}({\bf{y}}) = {\left\| {{\bf{y}} - {{\bf{x}}_c}{{\bf{\hat s}}_c}} \right\|_2^2}$
		\ENDFOR
		\STATE $id\left( {\bf{y}} \right) = \arg {\min _c}\left\{ {{e_c}} \right\}$
		\RETURN $id({\bf{y}})$
	\end{algorithmic}
\end{algorithm}

\subsection{Label Consistent K-SVD (LC-KSVD)}

\begin{algorithm}[!t]
	\scriptsize
	\caption{Label Consistent K-SVD}\label{Algorithm2}
	\hspace*{0.02in} {\bf Input:} ${\bf{X}}\in\mathbb{R}^{D\times N}$, ${\bf{H}}\in\mathbb{R}^{C\times N}$, ${\bf{Q}}\in\mathbb{R}^{K\times N}$, $\lambda$, $\omega$, $T$, $K$\\
	\hspace*{0.02in} {\bf Output:} ${\bf{B}}\in\mathbb{R}^{D\times K}$, ${\bf{W}}\in\mathbb{R}^{C\times K}$, ${\bf{A}}\in\mathbb{R}^{K\times K}$, ${\bf{S}}\in\mathbb{R}^{K\times N}$	
	\begin{algorithmic}[1]
		\STATE Compute ${{\bf{B}}_0}$ by combining class-specific dictionary items for each class using K-SVD~\cite{aharon2006k};
		\STATE Compute ${{\bf{S}}_0}$ for ${\bf{X}}$ and ${{\bf{B}}_0}$ using sparse coding;
		\STATE Compute ${{\bf{A}}_0}$ using ${\bf{A}} = {\bf{Q}}{{\bf{S}}^T}{\left( {{\bf{S}}{{\bf{S}}^T} + {\bf{I}}} \right)^{ - 1}}$;
		\STATE Cpmpute ${{\bf{W}}_0}$ using ${\bf{W}} = {\bf{H}}{{\bf{S}}^T}{\left( {{\bf{S}}{{\bf{S}}^T} + {\bf{I}}} \right)^{ - 1}}$;
		\STATE Solve Eq.(3); Use ${\left[ {\begin{array}{*{20}{c}}
				{{\bf{B}}_0}\\
				{\sqrt \omega  {{\bf{A}}_0}}\\
				{\sqrt \lambda  {{\bf{W}}_0}}
				\end{array}} \right]}$ to initialize the dictionary.
		\STATE Normalize ${\bf{B}}$ ${\bf{A}}$ ${\bf{W}}$:\\
		${\bf{B}} \leftarrow \left\{ {\frac{{{{\bf{b}}_1}}}{{{{\left\| {{{\bf{b}}_1}} \right\|}_2}}},\frac{{{{\bf{b}}_2}}}{{{{\left\| {{{\bf{b}}_2}} \right\|}_2}}}, \cdots ,\frac{{{{\bf{b}}_K}}}{{{{\left\| {{{\bf{b}}_K}} \right\|}_2}}}} \right\}$\\
		${\bf{A}} \leftarrow \left\{ {\frac{{{{\bf{a}}_1}}}{{{{\left\| {{{\bf{b}}_1}} \right\|}_2}}},\frac{{{{\bf{a}}_2}}}{{{{\left\| {{{\bf{b}}_2}} \right\|}_2}}}, \cdots ,\frac{{{{\bf{a}}_K}}}{{{{\left\| {{{\bf{b}}_K}} \right\|}_2}}}} \right\}$\\
		${\bf{W}} \leftarrow \left\{ {\frac{{{{\bf{w}}_1}}}{{{{\left\| {{{\bf{b}}_1}} \right\|}_2}}},\frac{{{{\bf{w}}_2}}}{{{{\left\| {{{\bf{b}}_2}} \right\|}_2}}}, \cdots ,\frac{{{{\bf{w}}_K}}}{{{{\left\| {{{\bf{b}}_K}} \right\|}_2}}}} \right\}$\\
		\RETURN  ${\bf{B}}$, ${\bf{W}}$, ${\bf{A}}$, ${\bf{S}}$
	\end{algorithmic}
\end{algorithm}

\cite{jiang2013label} proposed LC-KSVD to encourage the similarity among representations of samples belonging to the same class in D-KSVD. 
The authors proposed to combine the discriminative sparse codes error with the reconstruction error and the classification error to form a unified objective function, which gave discriminative sparse codes matrix ${\bf{Q}} = \left[ {{{\bf{q}}_1},{{\bf{q}}_2}, \cdots ,{{\bf{q}}_N}} \right] \in {{\mathbb{R}}^{K\times N}}$, label matrix ${\bf{H}} = \left[ {{{\bf{h}}_1},{{\bf{h}}_2}, \cdots ,{{\bf{h}}_N}} \right] \in {{\mathbb{R}}^{C\times N}}$ and training sample matrix ${\bf{X}}$. The objective function is defined as follows:
\begin{equation}
	\begin{split}
		< {\bf{B}},{\bf{W}},{\bf{A}},{\bf{S}} > &= \mathop {\arg \min }\limits_{{\bf{B}},{\bf{W}},{\bf{A}},{\bf{S}}} \left\| {{\bf{X}} - {\bf{BS}}} \right\|_F^2 + \lambda \left\| {{\bf{H}} - {\bf{WS}}} \right\|_F^2 \\&+ \omega \left\| {{\bf{Q}} - {\bf{AS}}} \right\|_F^2\\
		{\kern 12pt} & s.t.{\kern 3pt} {\left\| {{{\bf{s}}_i}} \right\|_0} < T {\kern 6pt} \left( {i = 1,2 \cdots ,N} \right)\\
		&{\rm{ = }}\mathop {{\rm{argmin}}}\limits_{{\bf{B}},{\bf{W}},{\bf{A}},{\bf{S}}} \left\| {\left[ {\begin{array}{*{20}{c}}
					{\bf{X}}\\
					{\sqrt \omega  {\bf{Q}}}\\
					{\sqrt \lambda  {\bf{H}}}
			\end{array}} \right] - \left[ {\begin{array}{*{20}{c}}
					{\bf{B}}\\
					{\sqrt \omega  {\bf{A}}}\\
					{\sqrt \lambda  {\bf{W}}}
			\end{array}} \right]{\bf{S}}} \right\|_F^2\\
		&s.t.{\kern 3pt}{\left\| {{{\bf{s}}_i}} \right\|_0} < T {\kern 6pt} \left( {i = 1,2 \cdots ,N} \right)
	\end{split}\label{LC-KSVD}
\end{equation}
where $T$ is the sparsity constraint factor, making sure that ${{{\bf{s}}_i}}$ has no more than $T$ nonzero entries. 
The dictionary ${\bf{B}} = \left[ {{{\bf{b}}_1},{{\bf{b}}_2}, \cdots ,{{\bf{b}}_K}} \right] \in {{\mathbb{R}}^{D\times K}}$, where $K>D$ is the number of atoms in the dictionary, and ${\bf{S}} = \left[ {{{\bf{s}}_1},{{\bf{s}}_2}, \cdots ,{{\bf{s}}_N}} \right] \in {{\mathbb{R}}^{K\times D}}$ is the sparse codes of training sample matrix ${\bf{X}}$. 
${\bf{W}} = \left[ {{{\bf{w}}_1},{{\bf{w}}_2}, \cdots ,{{\bf{w}}_K}} \right] \\ \in {{\mathbb{R}}^{C\times K}}$ is a classifier learned from the given label matrix ${\bf{H}}$. 
We hope the ${\bf{W}}$ can return the most probable class this sample belongs to. 
${\bf{A}} = \left[ {{{\bf{a}}_1},{{\bf{a}}_2}, \cdots ,{{\bf{a}}_K}} \right] \in {{\mathbb{R}}^{K\times K}}$ is a linear transformation relys on ${\bf{Q}}$. 
$\lambda$ and $\omega$ are the regularization parameters balancing the discriminative sparse codes error and the classification contribution to the overall objective, respectively. 
The algorithm is shown in Algorithm~\ref{Algorithm2}. 
Here, we denote $m\left( {m = 0,1,2, \cdots } \right)$ as the iteration number and ${\left(  \bullet  \right)_m}$ means the value of matrix $\left(  \bullet  \right)$ after ${m_{th}}$ iteration.

While the LC-KSVD algorithm exploits the $\ell_0$-norm regularization term to control the sparseness, it is difficult to find the optimal sparse solution to a general image recognition. The reason is that LC-KSVD use OMP method to optimise the objective function which usually obtain the suboptimal sparse solution unless finding the 
perfect initialized values.

\section{Methodology}
\label{Methodology}

In this section, we first give our proposed label embedded dictionary learning algorithm. Then we elaborate the optimization of the objective function. 

\subsection{Proposed Label Embedded Dictionary Learning (LEDL)}
\label{Methodology-A}

Motivated by that the optimal sparse solution can not be found easily with $\ell_0$-norm regularization term, we propose a novel dictionary learning method named label embedded dictionary learning (LEDL) for image classification.
This method introduces the $\ell_1$-norm regularization term to replace the $\ell_0$-norm regularization of LC-KSVD. 
Thus, we can freely select the basis vectors for linear fitting to get optimal sparse solution. 
The objection function is as follows:

\begin{equation}
	\begin{split}
		< {\bf{B}},{\bf{W}},{\bf{A}},{\bf{S}} > = &\mathop {\arg \min }\limits_{{\bf{B}},{\bf{W}},{\bf{A}},{\bf{S}}} \left\| {{\bf{X}} - {\bf{BS}}} \right\|_F^2 + \lambda \left\| {{\bf{H}} - {\bf{WS}}} \right\|_F^2 \\&+ \omega \left\| {{\bf{Q}} - {\bf{AS}}} \right\|_F^2 + 2\varepsilon {\left\| {\bf{S}} \right\|_{\ell_1}}\\
		{\rm{s}}.t.{\kern 4pt}\left\| {{{\bf{B}}_{ \bullet k}}} \right\|_2^2 \le 1, {\kern 1pt} {\kern 1pt} &\left\| {{{\bf{W}}_{ \bullet k}}} \right\|_2^2 \le 1,{\kern 1pt}
		\left\| {{{\bf{A}}_{ \bullet k}}} \right\|_2^2 \le 1{\kern 3pt}\left( {k = 1,2, \cdots K} \right)
	\end{split}\label{LEDL}
\end{equation}
where, ${\left(  \bullet  \right)_{ \bullet k}}$ denotes the $k_{th}$ column vector of matrix  $\left(  \bullet  \right)$. The $\ell_1$-norm regularization term is utilized to enforce sparsity and $\varepsilon$ is the regularization parameter which has the same function as $\alpha$ in Equation (\ref{SRC}).

\subsection{Optimization of Objective Function}
\label{Methodology-B}

\begin{figure*}
	\begin{center}
		\includegraphics[width = 1.0\linewidth]{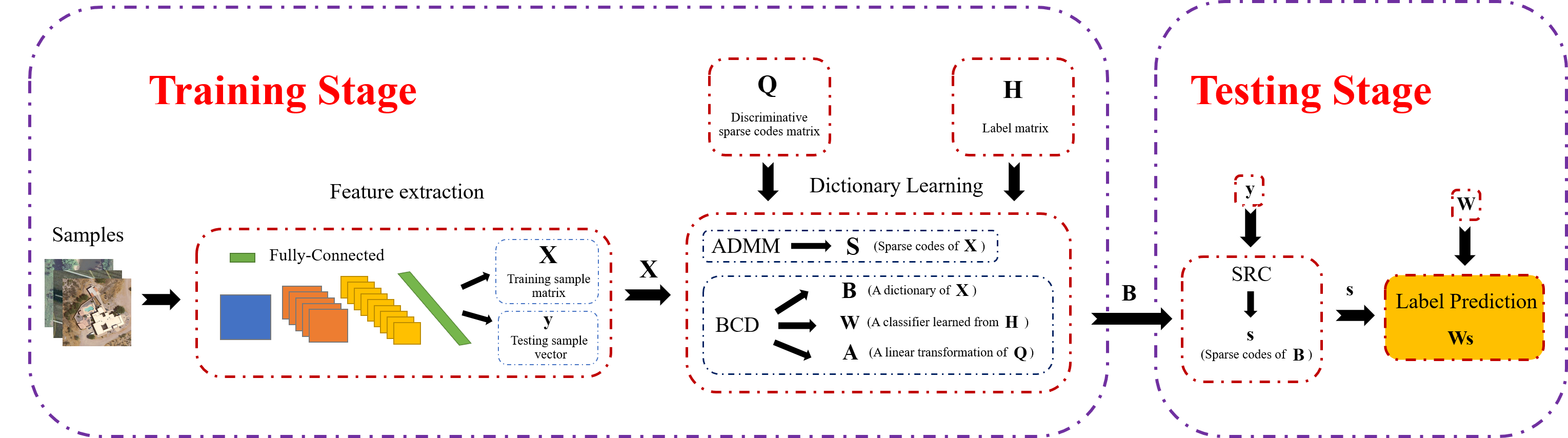}
	\end{center}
	\caption{The complete process of LEDL algorithm}
	\label{fig:Procedure4LEDL}
\end{figure*}

Consider the optimization problem (\ref{LEDL}) is not jointly convex in both ${\bf{S}}$, ${\bf{B}}$, ${\bf{W}}$ and ${\bf{A}}$, it is separately convex in either ${\bf{S}}$ (with ${\bf{B}}$, ${\bf{W}}$, ${\bf{A}}$ fixed), ${\bf{B}}$ (with ${\bf{S}}$, ${\bf{W}}$, ${\bf{A}}$ fixed), ${\bf{W}}$ (with ${\bf{S}}$, ${\bf{B}}$, ${\bf{A}}$ fixed) or ${\bf{A}}$ (with ${\bf{S}}$, ${\bf{B}}$, ${\bf{W}}$ fixed). 
To this end, the optimization problem can be recognised as four optimization subproblems which are finding sparse codes ($\bf{S}$) and learning bases ($\bf{B}$, $\bf{W}$, $\bf{A}$), respectively. Here, we employ the alternating direction method of multipliers (ADMM)~\cite{boyd2011distributed} framework to solve the first subproblem and the blockwise coordinate descent (BCD)~\cite{liu2014blockwise} algorithm for the rest subproblem. 
The complete process of LEDL is shown in Figure~\ref{fig:Procedure4LEDL}. 

\subsubsection{ADMM for finding sparse codes}

While fixing ${\bf{B}}$, ${\bf{W}}$ and ${\bf{A}}$, we introduce an auxiliary variable ${\bf{Z}}$ and reformulate the LEDL algorithm into a linear equality-constrained problem with respect to each iteration has the closed-form solution. The objective function is as follows:\\
\begin{equation}
	\begin{split}
		< {\bf{B}},{\bf{W}},{\bf{A}},{\bf{C}},{\bf{Z}} >  = &\mathop {\arg \min }\limits_{{\bf{B}},{\bf{W}},{\bf{A}},{\bf{C}},{\bf{Z}}} \left\| {{\bf{X}} - {\bf{BC}}} \right\|_F^2  + 2\varepsilon {\left\| {\bf{Z}} \right\|_{\ell_1}}
		\\&+ \lambda \left\| {{\bf{H}} - {\bf{WC}}} \right\|_F^2 + \omega \left\| {{\bf{Q}} - {\bf{AC}}} \right\|_F^2 
		\\s.t.{\kern 2pt} {\kern 2pt} {\bf{C}} = {\bf{Z}},{\kern 1pt} {\kern 1pt}
		&\left\| {{{\bf{B}}_{ \bullet k}}} \right\|_2^2 \le 1,{\kern 1pt} {\kern 1pt} {\kern 1pt} \left\| {{{\bf{W}}_{ \bullet k}}} \right\|_2^2 \le 1,{\kern 1pt} {\kern 1pt} {\kern 1pt}\\& \left\| {{{\bf{A}}_{ \bullet k}}} \right\|_2^2 \le 1(k = 1,2 \cdots K)
	\end{split}\label{LEDL_ADMM1}
\end{equation}

While utilising the ADMM framework with fixed ${\bf{B}}$, ${\bf{W}}$ and ${\bf{A}}$, the lagrangian function of the problem (\ref{LEDL_ADMM1}) is rewritten as:
\begin{equation}
	\begin{split}
		<{\bf{C}},{\bf{Z}},{\bf{L}}  >  = &\mathop {\arg \min }\limits_{{\bf{C}},{\bf{Z}},{\bf{L}} } \left\| {{\bf{X}} - {\bf{BC}}} \right\|_F^2 + \lambda \left\| {{\bf{H}} - {\bf{WC}}} \right\|_F^2 
		+ \omega \left\| {{\bf{Q}} - {\bf{AC}}} \right\|_F^2
		\\&+ 2\varepsilon {\left\| {\bf{Z}} \right\|_{\ell_1}} 
		+ 2{{\bf{L}}^T}({\bf{C}} - {\bf{Z}}) 
		+ \rho \left\| {{\bf{C}} - {\bf{Z}}} \right\|_F^2\\
	\end{split}\label{LEDL_ADMM2}
\end{equation}
where ${\bf{L}} = \left[ {{{\bf{l}}_1},{{\bf{l}}_2}, \cdots ,{{\bf{l}}_N}} \right] \in {{\mathbb{R}}^{K\times N}}$ is the augmented lagrangian multiplier and $ \rho>0$ is the penalty parameter. 
After fixing ${\bf{B}}$, ${\bf{W}}$ and ${\bf{A}}$, we initialize the ${{\bf{C}}_0}$, ${{\bf{Z}}_0}$ and ${{\bf{L}}_0}$ to be zero matrices. 
Equation (\ref{LEDL_ADMM2}) can be solved as follows:\\\\
$\left( 1 \right)$ { Updating ${{\bf{C}}}$ while fixing ${{\bf{Z}}}$, ${{\bf{L}}}$, ${{\bf{B}}}$, ${{\bf{W}}}$ and ${{\bf{A}}}$}:
\begin{equation}
	\begin{split}
		{{\bf{C}}_{m + 1}}  = < {{\bf{B}}_m},{{\bf{W}}_m},{{\bf{A}}_m},{{\bf{C}}_m},{{\bf{Z}}_m},{{\bf{L}}_m} > 
	\end{split}\label{Update_C1}
\end{equation}
The closed form solution of ${\bf{C}}$ is
\begin{equation}
	\begin{split}
		{{\bf{C}}_{m + 1}}& = {\left( {{{\bf{B}}_m}^T{{\bf{B}}_m} + \lambda {{\bf{W}}_m}^T{{\bf{W}}_m} + \omega {{\bf{A}}_m}^T{{\bf{A}}_m} + \rho {\bf{I}}} \right)^{ - 1}} \\& \times \left( {{{\bf{B}}_m}^T{{\bf{X}}} + \lambda {{\bf{W}}_m}^T{{\bf{H}}} + \omega {{\bf{A}}_m}^T{{\bf{Q}}} + \rho {{\bf{Z}}_m} - {\bf{L}}_m} \right)
	\end{split}\label{Update_C2}
\end{equation}
$\left( 2 \right)$ { Updating ${{\bf{Z}}}$ while fixing ${{\bf{C}}}$, ${{\bf{L}}}$, ${{\bf{B}}}$, ${{\bf{W}}}$ and ${{\bf{A}}}$}\\
\begin{equation}
	\begin{split}
		{{\bf{Z}}_{m + 1}}  = < {{\bf{B}}_m},{{\bf{W}}_m},{{\bf{A}}_m},{{\bf{C}}_{m+1}},{{\bf{Z}}_m},{{\bf{L}}_m} > 
	\end{split}\label{Update_Z1}
\end{equation}
The closed form solution of {\bf{Z}} is
\begin{equation}
	\begin{split}
		{{\bf{Z}}_{m + 1}} = \max \left\{ {{{\bf{C}}_{m + 1}} + \frac{{{{\bf{L}}_m}}}{\rho } - \frac{\varepsilon }{\rho }{\bf{I}},{\bf{0}}} \right\}\\
		+ \min \left\{ {{{\bf{C}}_{m + 1}} + \frac{{{{\bf{L}}_m}}}{\rho } + \frac{\varepsilon }{\rho }{\bf{I}},{\bf{0}}} \right\}
	\end{split}\label{Update_Z2}
\end{equation}
where $\bf{I}$ is the identity matrix and $\bf{0}$ is the zero matrix.
\\$\left( 3 \right)$ { Updating the Lagrangian multiplier} ${{\bf{L}}}$\\
\begin{equation}
	\begin{split}
		{{\bf{L}}_{m + 1}} = {{\bf{L}}_m} + \rho \left( {{{\bf{C}}_{m + 1}} - {{\bf{Z}}_{m + 1}}} \right)
	\end{split}\label{Update_L1}
\end{equation}
where the $\rho$ in Equation (\ref{Update_L1}) is the gradient of gradient descent (GD) method, which has no relationship with the  $\rho$ in Equation (\ref{LEDL_ADMM2}). In order to make better use of ADMM framework, the $\rho$ in Equation (\ref{Update_L1}) can be rewritten as $\theta$.
\begin{equation}
	\begin{split}
		{{\bf{L}}_{m + 1}} = {{\bf{L}}_m} + \theta \left( {{{\bf{C}}_{m + 1}} - {{\bf{Z}}_{m + 1}}} \right)
	\end{split}\label{Update_L2}
\end{equation}

\subsubsection{BCD for learning bases}
Without consisdering the sparseness regulariation term in Equation (\ref{LEDL_ADMM1}), the constrained minimization problem of (\ref{LEDL}) with respect to the single column has the closed-form solution which can be solved by BCD method. The objective function can be rewritten as follows:\\
\begin{equation}
	\begin{split}
		<{\bf{B}},{\bf{W}},{\bf{A}}  >  &= \mathop {\arg \min }\limits_{{\bf{B}},{\bf{W}},{\bf{A}} } \left\| {{\bf{X}} - {\bf{BC}}} \right\|_F^2 + \lambda \left\| {{\bf{H}} - {\bf{WC}}} \right\|_F^2 
		+ \omega \left\| {{\bf{Q}} - {\bf{AC}}} \right\|_F^2
		\\&+ 2\varepsilon {\left\| {\bf{Z}} \right\|_{\ell_1}} 
		+ 2{{\bf{L}}^T}({\bf{C}} - {\bf{Z}}) + \rho \left\| {{\bf{C}} - {\bf{Z}}} \right\|_F^2\\
		s.t.{\kern 2pt} {\kern 2pt} {\kern 1pt} {\kern 1pt}
		&\left\| {{{\bf{B}}_{ \bullet k}}} \right\|_2^2 \le 1,{\kern 1pt} {\kern 1pt} {\kern 1pt} \left\| {{{\bf{W}}_{ \bullet k}}} \right\|_2^2 \le 1,{\kern 1pt} {\kern 1pt} {\kern 1pt} 
		\left\| {{{\bf{A}}_{ \bullet k}}} \right\|_2^2 \le 1(k = 1,2 \cdots K)
	\end{split}\label{LEDL_BCD}
\end{equation}
We initialize ${{\bf{B}}_0}$, ${{\bf{W}}_0}$ and  ${{\bf{A}}_0}$ to be random matrices and normalize them, respectively. After that we use BCD method to update ${{\bf{B}}}$, ${{\bf{W}}}$ and ${{\bf{A}}}$.\\\\
$\left( 1 \right)$ { Updating ${{\bf{B}}}$ while fixing ${{\bf{C}}}$, ${{\bf{L}}}$, ${{\bf{Z}}}$, ${{\bf{W}}}$ and ${{\bf{A}}}$}\\
\begin{equation}
	\begin{split}
		{{\bf{B}}_{m + 1}}  = < {{\bf{B}}_m},{{\bf{W}}_m},{{\bf{A}}_m},{{\bf{C}}_{m+1}},{{\bf{Z}}_{m+1}},{{\bf{L}}_{m+1}} > 
	\end{split}\label{Update_B1}
\end{equation}
The closed-form solution of single column of ${{\bf{B}}}$ is
\begin{equation}
	\begin{split}
		&\left( {{{\bf{B}}_{ \bullet k}}} \right){{\kern 1pt} _{m + 1}} 
		= \frac{{{\bf{X}}{{\left[ {{{\left( {{{\bf{C}}_{k \bullet }}} \right)}_{m + 1}}} \right]}^T} - {{\left( {{{{\bf{\tilde B}}}^k}} \right)}_m}{{\bf{C}}_{m + 1}}{{\left[ {{{\left( {{{\bf{C}}_{k \bullet }}} \right)}_{m + 1}}} \right]}^T}}}{{{{\left\| {{\bf{X}}{{\left[ {{{\left( {{{\bf{C}}_{k \bullet }}} \right)}_{m + 1}}} \right]}^T} - {{\left( {{{{\bf{\tilde B}}}^k}} \right)}_m}{{\bf{C}}_{m + 1}}{{\left[ {{{\left( {{{\bf{C}}_{k \bullet }}} \right)}_{m + 1}}} \right]}^T}} \right\|}_2}}}{\kern 1pt}
	\end{split}\label{Update_B2}
\end{equation}
where ${\bf{\tilde B}}^k = \left\{ {\begin{array}{*{20}{c}}
	{{{\bf{B}}_{ \bullet p}},p \ne k}\\
	{{\bf{0}},{\kern 1pt} {\kern 1pt} {\kern 1pt} {\kern 1pt} {\kern 1pt} {\kern 1pt} {\kern 1pt} {\kern 1pt} {\kern 1pt} {\kern 1pt} {\kern 1pt} p = k}
	\end{array}} \right.$, ${\left(  \bullet  \right)_{k \bullet }}$ denotes the $k_{th}$ row vector of matrix $\left(  \bullet  \right)$.\\\\
$\left( 2 \right)$ { Updating ${{\bf{W}}}$ while fixing ${{\bf{C}}}$, ${{\bf{L}}}$, ${{\bf{Z}}}$, ${{\bf{B}}}$ and ${{\bf{A}}}$}\\
\begin{equation}
	\begin{split}
		{{\bf{W}}_{m + 1}}  = < {{\bf{B}}_{m+1}},{{\bf{W}}_m},{{\bf{A}}_m},{{\bf{C}}_{m+1}},{{\bf{Z}}_{m+1}},{{\bf{L}}_{m+1}} > 
	\end{split}\label{Update_W1}
\end{equation}
The closed-form solution of single column of ${{\bf{W}}}$ is
\begin{equation}
	\begin{split}
		&\left( {{{\bf{W}}_{ \bullet k}}} \right){{\kern 1pt} _{m + 1}} 
		= \frac{{{\bf{H}}{{\left[ {{{\left( {{{\bf{C}}_{k \bullet }}} \right)}_{m + 1}}} \right]}^T} - {{\left( {{{{\bf{\tilde W}}}^k}} \right)}_m}{{\bf{C}}_{m + 1}}{{\left[ {{{\left( {{{\bf{C}}_{k \bullet }}} \right)}_{m + 1}}} \right]}^T}}}{{{{\left\| {{\bf{H}}{{\left[ {{{\left( {{{\bf{C}}_{k \bullet }}} \right)}_{m + 1}}} \right]}^T} - {{\left( {{{{\bf{\tilde W}}}^k}} \right)}_m}{{\bf{C}}_{m + 1}}{{\left[ {{{\left( {{{\bf{C}}_{k \bullet }}} \right)}_{m + 1}}} \right]}^T}} \right\|}_2}}}{\kern 1pt} 
	\end{split}\label{Update_W2}
\end{equation}
where ${\bf{\tilde W}}^k = \left\{ {\begin{array}{*{20}{c}}
	{{{\bf{W}}_{ \bullet p}},p \ne k}\\
	{{\bf{0}},{\kern 1pt} {\kern 1pt} {\kern 1pt} {\kern 1pt} {\kern 1pt} {\kern 1pt} {\kern 1pt} {\kern 1pt} {\kern 1pt} {\kern 1pt} {\kern 1pt} p = k}
	\end{array}} \right.$.\\\\
$\left( 3 \right)$ { Updating ${{\bf{A}}}$ while fixing ${{\bf{C}}}$, ${{\bf{L}}}$, ${{\bf{Z}}}$, ${{\bf{B}}}$ and ${{\bf{W}}}$}\\
\begin{equation}
	\begin{split}
		{{\bf{A}}_{m + 1}}  = < {{\bf{B}}_{m+1}},{{\bf{W}}_{m+1}},{{\bf{A}}_m},{{\bf{C}}_{m+1}},{{\bf{Z}}_{m+1}},{{\bf{L}}_{m+1}} > 
	\end{split}\label{Update_A1}
\end{equation}
The closed-form solution of single column of ${{\bf{A}}}$ is
\begin{equation}
	\begin{split}
		&\left( {{{\bf{A}}_{ \bullet k}}} \right){{\kern 1pt} _{m + 1}} 
		= \frac{{{\bf{Q}}{{\left[ {{{\left( {{{\bf{C}}_{k \bullet }}} \right)}_{m + 1}}} \right]}^T} - {{\left( {{{{\bf{\tilde A}}}^k}} \right)}_m}{{\bf{C}}_{m + 1}}{{\left[ {{{\left( {{{\bf{C}}_{k \bullet }}} \right)}_{m + 1}}} \right]}^T}}}{{{{\left\| {{\bf{Q}}{{\left[ {{{\left( {{{\bf{C}}_{k \bullet }}} \right)}_{m + 1}}} \right]}^T} - {{\left( {{{{\bf{\tilde A}}}^k}} \right)}_m}{{\bf{C}}_{m + 1}}{{\left[ {{{\left( {{{\bf{C}}_{k \bullet }}} \right)}_{m + 1}}} \right]}^T}} \right\|}_2}}}{\kern 1pt} 
	\end{split}\label{Update_A2}
\end{equation}
where ${\bf{\tilde A}}^k = \left\{ {\begin{array}{*{20}{c}}
	{{{\bf{A}}_{ \bullet p}},p \ne k}\\
	{{\bf{0}},{\kern 1pt} {\kern 1pt} {\kern 1pt} {\kern 1pt} {\kern 1pt} {\kern 1pt} {\kern 1pt} {\kern 1pt} {\kern 1pt} {\kern 1pt} {\kern 1pt} p = k}
	\end{array}} \right.$.\\

\begin{figure*}
	\begin{center}
		\includegraphics[width=0.8\linewidth]{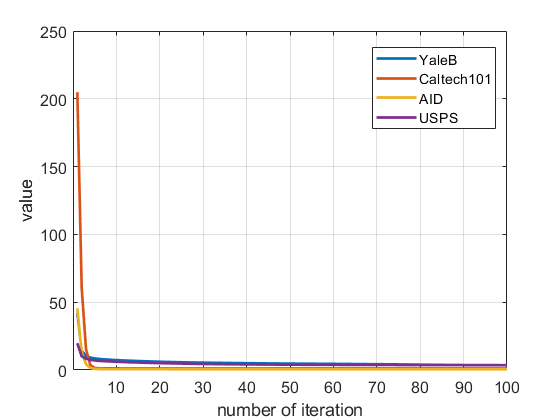}
	\end{center}
	\caption{
		Convergence curve of LEDL Algorithm on four datasets.
	}
	\label{fig:Convergence}
\end{figure*}

\subsubsection{Convergence Analysis}
Assume that the result of the objective function after ${m_{th}}$ iteration is defined as $f\left({{{\bf{C}}_m},{{\bf{Z}}_m},{{\bf{L}}_m},{{\bf{B}}_m},{{\bf{W}}_m},{{\bf{A}}_m}} \right)$. 
Since the minimum point is obtained by ADMM and BCD methods, each method will monotonically decrease the corresponding objective function after about 100 iterations. Considering that the objective function is obviously bounded below and satisfies the Equation (\ref{converge}), it converges. 
Figure~\ref{fig:Convergence} shows the convergence curve of the proposed LEDL algorithm by using four well-known datasets. The results demonstrate that our proposed LEDL algorithm has fast convergence and low complexity.

\begin{equation}
	\begin{split}
		&f\left({{{\bf{C}}_m},{{\bf{Z}}_m},{{\bf{L}}_m},{{\bf{B}}_m},{{\bf{W}}_m},{{\bf{A}}_m}} \right) 
		\\\ge &f\left( {{{\bf{C}}_{m + 1}},{{\bf{Z}}_{m + 1}},{{\bf{L}}_{m + 1}},{{\bf{B}}_m},{{\bf{W}}_m},{{\bf{A}}_m}} \right)
		\\\ge &f\left( {{{\bf{C}}_{m + 1}},{{\bf{Z}}_{m + 1}},{{\bf{L}}_{m + 1}},{{\bf{B}}_{m + 1}},{{\bf{W}}_{m + 1}},{{\bf{A}}_{m + 1}}} \right)
	\end{split}\label{converge}
\end{equation}

\subsubsection{Overall Algorithm}
The overall updating procedures of proposed LEDL algorithm is summarized in Algorithm~\ref{Algorithm3}. Here, $maxiter$ is the maximum number of iterations, ${\bf{1}}\in\mathbb{R}^{K\times K}$ is a squre matrix with all elements 1 and $ \odot $ indicates element dot product. By iterating ${\bf{C}}$, ${\bf{Z}}$, ${\bf{L}}$, ${\bf{B}}$, ${\bf{W}}$ and ${\bf{A}}$ alternately, the sparse codes are obtained, and the corresponding bases are learned.
\begin{algorithm}[!ht]
	\scriptsize
	\caption{Label Embedded Dictionary Learning}\label{Algorithm3}
	\hspace*{0.02in} {\bf Input:} ${\bf{X}}\in\mathbb{R}^{D\times N}$, ${\bf{H}}\in\mathbb{R}^{C\times N}$, ${\bf{Q}}\in\mathbb{R}^{K\times N}$, $\lambda$, $\omega$, $\varepsilon$, $\rho$, $\theta$, $K$\\
	\hspace*{0.02in} {\bf Output:} ${\bf{B}}\in\mathbb{R}^{D\times K}$, ${\bf{W}}\in\mathbb{R}^{C\times K}$, ${\bf{A}}\in\mathbb{R}^{K\times K}$, ${\bf{C}}\in\mathbb{R}^{K\times N}$
	
	\begin{algorithmic}[1]
		\STATE ${{\bf{C}}_0} \leftarrow zeros\left( {K,N} \right)$, ${{\bf{Z}}_0} \leftarrow zeros\left( {K,N} \right)$, 
		${{\bf{L}}_0} \leftarrow zeros\left( {K,N} \right)$
		\STATE ${{\bf{B}}_0} \leftarrow rand\left( {D,K} \right)$, 
		${{\bf{W}}_0} \leftarrow rand\left( {C,K} \right)$,
		${{\bf{A}}_0} \leftarrow rand\left( {K,K} \right)$
		\STATE ${{\bf{B}}_{ \bullet k}} = \frac{{{{\bf{B}}_{ \bullet k}}}}{{{{\left\| {{{\bf{B}}_{ \bullet k}}} \right\|}_2}}}$, ${{\bf{W}}_{ \bullet k}} = \frac{{{{\bf{W}}_{ \bullet k}}}}{{{{\left\| {{{\bf{W}}_{ \bullet k}}} \right\|}_2}}}$, ${{\bf{A}}_{ \bullet k}} = \frac{{{{\bf{A}}_{ \bullet k}}}}{{{{\left\| {{{\bf{A}}_{ \bullet k}}} \right\|}_2}}}$,	$(k = 1,2 \cdots K)$
		\STATE $m = 0$
		\WHILE {$m \le \max iter$}
		\STATE $m \leftarrow m + 1$
		\STATE \textbf{Update ${\bf{C}}$:}\\
		\STATE${{\bf{C}}_{m + 1}} = {\left( {{{\bf{B}}_m}^T{{\bf{B}}_m} + \lambda {{\bf{W}}_m}^T{{\bf{W}}_m} + \omega {{\bf{A}}_m}^T{{\bf{A}}_m} + \rho {\bf{I}}} \right)^{ - 1}}$
		\STATE $ {\kern 22pt} \times \left( {{{\bf{B}}_m}^T{{\bf{X}}} + \lambda {{\bf{W}}_m}^T{{\bf{H}}} + \omega {{\bf{A}}_m}^T{{\bf{Q}}} + \rho {{\bf{Z}}_m} - {{\bf{L}}_m}} \right)$
		\STATE \textbf{Update ${\bf{Z}}$:}\\
		\STATE ${{\bf{Z}}_{m + 1}} = \max \left\{ {{{\bf{C}}_{m + 1}} + \frac{{{{\bf{L}}_m}}}{\rho } - \frac{\varepsilon }{\rho }{\bf{I}},{\bf{0}}} \right\}$ 
		\STATE ${\kern 22pt}+ \min \left\{ {{{\bf{C}}_{m + 1}} + \frac{{{{\bf{L}}_m}}}{\rho } + \frac{\varepsilon }{\rho }{\bf{I}},{\bf{0}}} \right\}$
		\STATE \textbf{Update ${\bf{L}}$:}\\
		\STATE ${{\bf{L}}_{m + 1}} = {{\bf{L}}_m} + \theta \left( {{{\bf{C}}_{m + 1}} - {{\bf{Z}}_{m + 1}}} \right)$
		\STATE \textbf{Update ${\bf{B}}$, ${\bf{W}}$, ${\bf{A}}$:}\\
		\STATE 	Compute ${{\bf{D}}_{m + 1}} = \left( {{{\bf{C}}_{m + 1}}{{\bf{C}}_{m + 1}}^T} \right) \odot \left( {{\bf{1}} - {\bf{I}}} \right)$	
		\FOR{$\scriptsize k=1$;$\scriptsize k\le \scriptsize K$;$\scriptsize k\!+\!+$}
		\STATE $\left( {{{\bf{B}}_{ \bullet k}}} \right){{\kern 1pt} _{m + 1}} = \frac{{{\bf{X}}{{\left[ {{{\left( {{{\bf{C}}_{k \bullet }}} \right)}_{m + 1}}} \right]}^T} - {{\bf{B}}_m}{{\left( {{{\bf{D}}_{ \bullet k}}} \right)}_{m + 1}}}}{{{{\left\| {{\bf{X}}{{\left[ {{{\left( {{{\bf{C}}_{k \bullet }}} \right)}_{m + 1}}} \right]}^T} - {{\bf{B}}_m}{{\left( {{{\bf{D}}_{ \bullet k}}} \right)}_{m + 1}}} \right\|}_2}}}{\kern 1pt}$
		\STATE $\left( {{{\bf{W}}_{ \bullet k}}} \right){{\kern 1pt} _{m + 1}} = \frac{{{\bf{H}}{{\left[ {{{\left( {{{\bf{C}}_{k \bullet }}} \right)}_{m + 1}}} \right]}^T} - {{\bf{W}}_m}{{\left( {{{\bf{D}}_{ \bullet k}}} \right)}_{m + 1}}}}{{{{\left\| {{\bf{H}}{{\left[ {{{\left( {{{\bf{C}}_{k \bullet }}} \right)}_{m + 1}}} \right]}^T} - {{\bf{W}}_m}{{\left( {{{\bf{D}}_{ \bullet k}}} \right)}_{m + 1}}} \right\|}_2}}}{\kern 1pt} $
		\STATE $\left( {{{\bf{A}}_{ \bullet k}}} \right){{\kern 1pt} _{m + 1}} = \frac{{{\bf{Q}}{{\left[ {{{\left( {{{\bf{C}}_{k \bullet }}} \right)}_{m + 1}}} \right]}^T} - {{\bf{A}}_m}{{\left( {{{\bf{D}}_{ \bullet k}}} \right)}_{m + 1}}}}{{{{\left\| {{\bf{Q}}{{\left[ {{{\left( {{{\bf{C}}_{k \bullet }}} \right)}_{m + 1}}} \right]}^T} - {{\bf{A}}_m}{{\left( {{{\bf{D}}_{ \bullet k}}} \right)}_{m + 1}}} \right\|}_2}}}{\kern 1pt} $
		\ENDFOR
		\STATE \textbf{Update the objective function:}\\
		\STATE $f = \left\| {{\bf{X}} - {\bf{BC}}} \right\|_F^2 + \lambda \left\| {{\bf{Y}} - {\bf{WC}}} \right\|_F^2 + \omega \left\| {{\bf{Q}} - {\bf{AC}}} \right\|_F^2 + 2\varepsilon {\left\| {\bf{Z}} \right\|_{\ell_1}} + {{\bf{L}}^T}({\bf{C}} - {\bf{Z}}) + \rho \left\| {{\bf{C}} - {\bf{Z}}} \right\|_F^2$ 
		\ENDWHILE
		\RETURN ${\bf{B}}$, ${\bf{W}}$, ${\bf{A}}$, ${\bf{C}}$
	\end{algorithmic}
\end{algorithm}

In testing stage, the constraint terms are based on $\ell_1$-norm sparse constraint. Here, we exploit the learned dictionary ${\bf{D}}$ to fit the testing sample $\bf{y}$ to obtain the sparse codes ${\bf{s}}$. Then, we use the trained classfier ${\bf{W}}$ to predict the label of ${\bf{y}}$ by calculating $\max \left\{ {{\bf{Ws}}} \right\}$.

\section{Experimental results}
\label{Experimental results}

In this section, we utilize several datasets (Extended YaleB~\cite{georghiades2001few}, CMU PIE~\cite{sim2002cmu}, UC Merced Land Use~\cite{yang2010bag}, AID~\cite{xia2017aid}, Caltech101~\cite{fei2007learning} and USPS~\cite{hull1994database}) to evaluate the performance of our algorithm and compare it with other state-of-the-art methods such as SRC~\cite{wright2009robust}, LC-KSVD~\cite{jiang2013label}, CRC~\cite{zhang2011sparse} and CSDL-SRC~\cite{liu2016face}. In the following subsection, we first give the experimental settings. Then experiments on these six datasets are analyzed. Moreover, some discussions are listed finally.

\begin{table}
	\scriptsize
	\caption{Classification rates ($\%$) on different datasets}
	\label{table1}
	\begin{center}
		\begin{tabular}{cccccc}
			\multicolumn{1}{c}{\bf Datasets$\backslash$Methods}  &\multicolumn{1}{c}{\bf SRC } &\multicolumn{1}{c}{\bf CRC}
			&\multicolumn{1}{c}{\bf CSDL-SRC} 
			&\multicolumn{1}{c}{\bf LC-KSVD}
			&\multicolumn{1}{c}{\bf LEDL} 
			\\ \hline
			Extended YaleB    &$79.1$   &$79.2$   &$80.2$   &$73.5$  &$\bf81.3$\\
			CMU PIE           &$73.7$   &$73.3$   &$77.4$   &$67.1$     &$\bf77.7$\\
			UC-Merced         &$80.4$   &$80.7$   &$80.5$   &$79.4$     &$\bf80.7$\\
			AID               &$71.6$   &$72.6$   &$71.6$   &$70.2$     &$\bf72.9$\\
			Caltech101        &$89.4$   &$89.4$   &$89.4$   &$88.3$     &$\bf90.1$\\
			USPS              &$78.4$   &$77.9$   &$78.8$   &$71.1$     &$\bf81.1$\\
			\hline\\
		\end{tabular}
	\end{center}
\end{table}

\subsection{Experimental settings}

For all the datasets, in order to eliminate the randomness, we carry out every experiment 8 times and the mean of the classification rates is reported. And we randomly select 5 samples per class for training in all the experiments.
For Extended YaleB dataset and CMU PIE dataset, each image is cropped to $32 \times 32$, pulled into column vector, and $\ell_2$ normalized to form the raw $\ell_2$ normalized features. For UC Merced Land Use dataset, AID dataset, we use resnet model~\cite{he2016deep} to extract the features. Specifically, the layer $pool5$ is utilized to extract 2048-dimensional vectors for them. For Caltech101 dataset, we use the layer $pool5$ of resnet model and spatial pyramid matching (SPM) with two layers (the second layer include five part, such as left upper, right upper, left lower, right lower, center) to extract 12288-dimensional vectors. And finally, each of the images in USPS dataset is resized into $16 \times 16$ vectors.

For convenience, the dictionary size ($K$) is fixed to the twice the number of training samples. In addition, we set $\rho = 1$ and initial $\theta = 0.5$, then decrease the $\theta$ in each iteration.
Moreover, there are other three parameters ($\lambda$, $\omega$ and $\varepsilon$) need to be adjust to achieve the highst classification rates. The details are showed in the following subsections.

\subsection{Extended YaleB Dataset}
\begin{figure}[h!]
	\begin{center}
	\includegraphics[width = 0.8\linewidth]{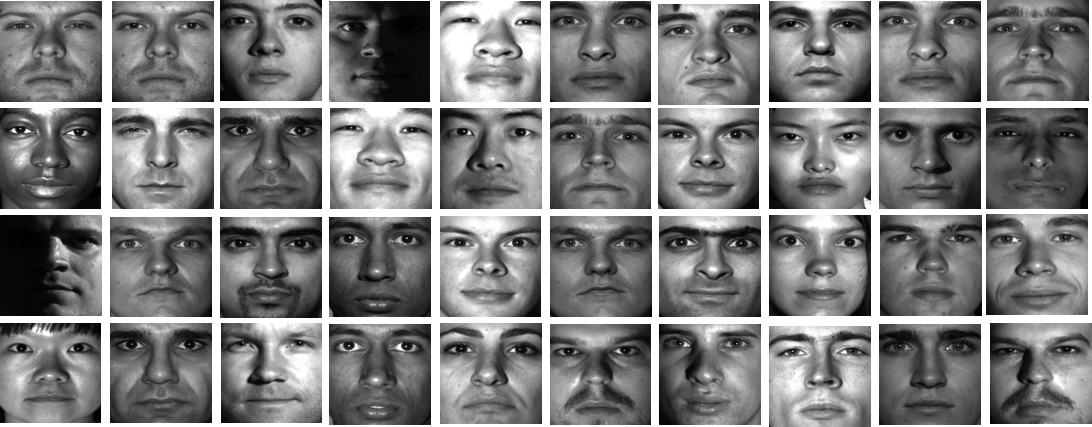}
	\end{center}
	\caption{Examples of the Extended YaleB dataset}
	\label{fig:YaleB}
\end{figure}

The Extended YaleB dataset contains $2{,}432$ face images from 38 individuals, each having 64 frontal images under varying illumination conditions. Figure~\ref{fig:YaleB} shows some images of the dataset.

\begin{figure*}
	\begin{center}
		\includegraphics[width=1.0\linewidth]{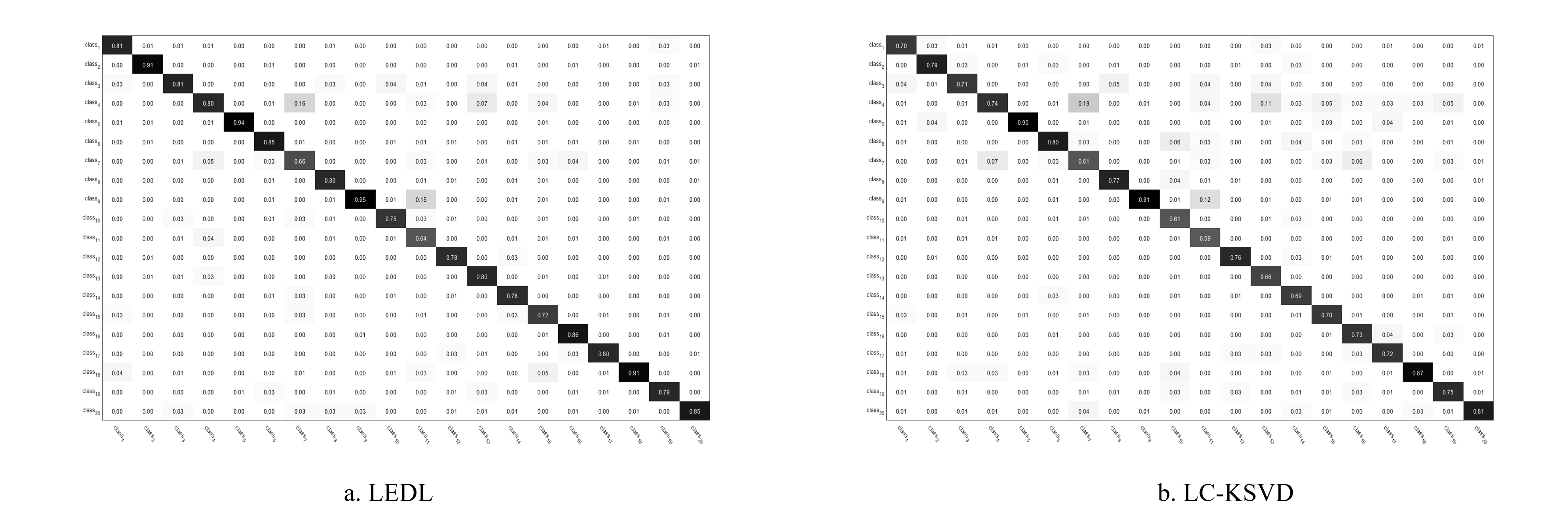}
	\end{center}
	\caption{Confusion matrices on Extended YaleB dataset}
	\label{fig:ConfusionMatrix4ExtendedYaleB}
\end{figure*}
In addition, we set $\lambda = {2^{ - 3}}$, $\omega = {2^{ - 11}}$, $\varepsilon = {2^{ - 8}}$ in our experiment. The experimental results are summarized in Table (\ref{table1}). We can see that our proposed LEDL algorithm achieves superior performance to other classical classification methods by an improvement of at least $1.1$$\%$. 
Compared with $\ell_0$-norm sparsity constraint based dictionary learning algorithm LC-KSVD, our proposed $\ell_1$-norm sparsity constraint based dictionary learning algorithm LEDL algorithm exceeds it $7.8$$\%$. The reason of the high improvement between LC-KSVD and LEDL is that $\ell_0$-norm sparsity constraint leads to NP-hard problem which is not conductive to finding the optimal sparse solution for the dictionary.
In order to further illustrate the performance of our method, we choose the first 20 classes samples as a subdataset and show the confusion matrices in Figure \ref{fig:ConfusionMatrix4ExtendedYaleB}. As can be seen that, our method achieves higher classification rates in all the chosen $20$ classes than LC-KSVD. Especially in class1, class2, class3, class10, class16, LEDL can achieve at least $10.0$$\%$ performance gain than LC-KSVD.

\subsection{CMU PIE Dataset}
\begin{figure}[h!]
	\begin{center}
		\includegraphics[width = 0.8\linewidth]{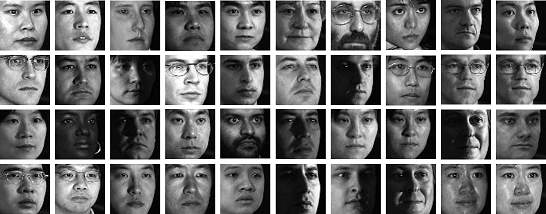}
	\end{center}
	\caption{Examples of the CMU PIE dataset}
	\label{fig:CMU_PIE}
\end{figure}

The CMU PIE dataset consists of $41{,}368$ images of 68 individuals with 43 different illumination conditions. Each human is under 13 different poses and with 4 different expressions. In Figure \ref{fig:CMU_PIE}, we list several samples from this dataset.

The comparasion results are showed in Table \ref{table1}, we can see that our proposed LEDL algorithm outperforms over other well-known methods by an improvement of at least $0.5$$\%$. To be attention, LEDL is capable of exceeding LC-KSVD $10.6$$\%$ in this dataset. The optimal parameters are $\lambda = {2^{ - 3}}$, $\omega = {2^{ - 11}}$, $\varepsilon = {2^{ - 8}}$. 

\subsection{UC Merced Land Use Dataset}
\begin{figure}[h!]
	\begin{center}
		\includegraphics[width = 0.8\linewidth]{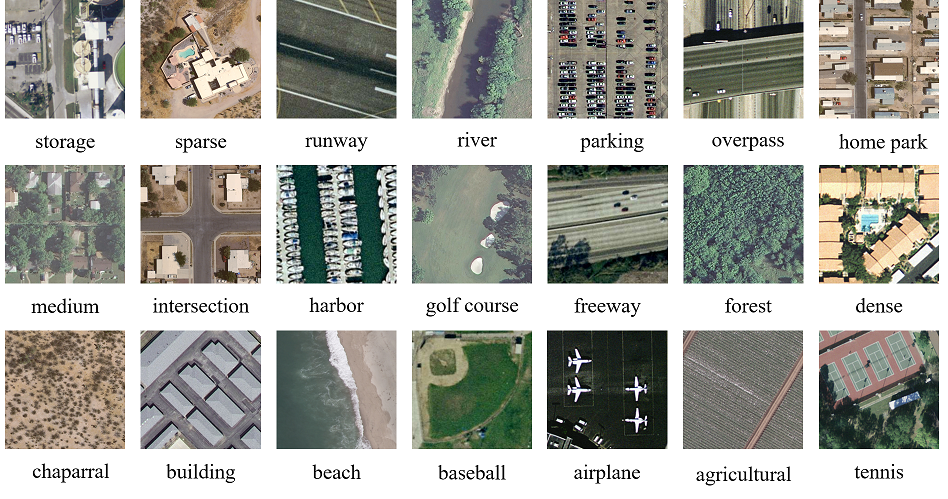}
	\end{center}
	\caption{Examples of the UC Merced dataset}\label{fig:UCMerced}
\end{figure}

The UC Merced Land Use dataset is widely used for aerial image classification. It consists of totally $2{,}100$ land-use images of $21$ classes. Some samples are showed in Figure \ref{fig:UCMerced}.

\begin{figure*}
	\begin{center}
		\includegraphics[width=1.0\linewidth]{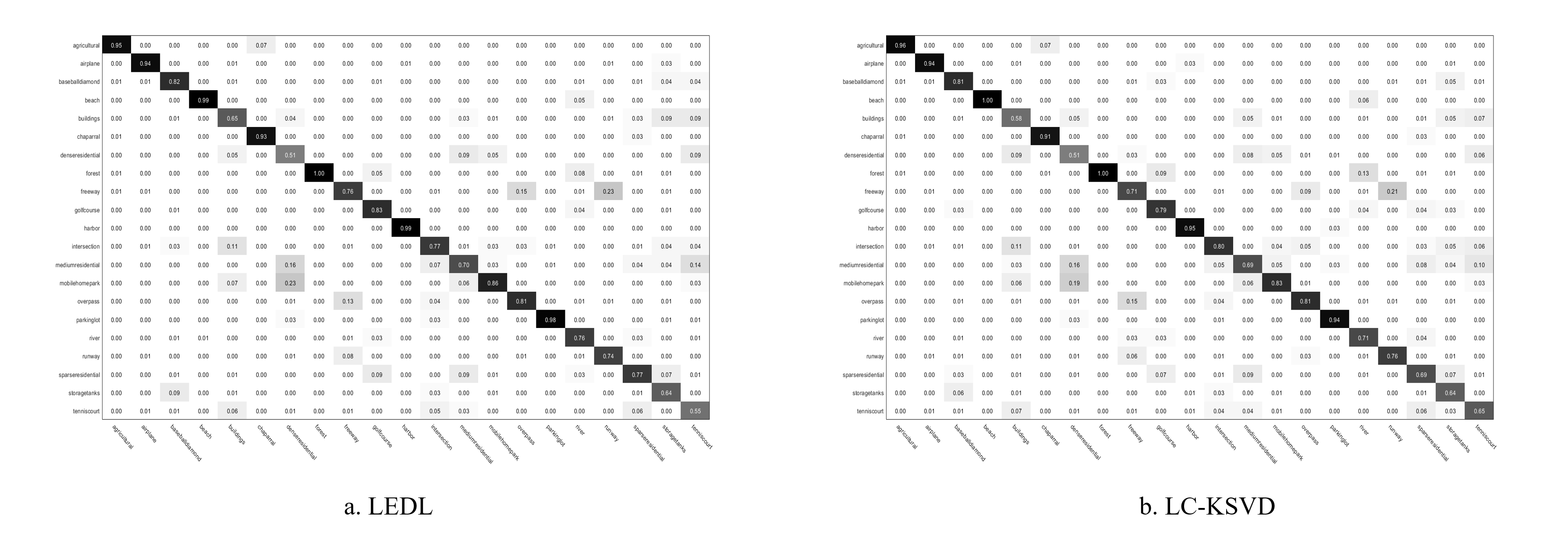}
	\end{center}
	\caption{Confusion matrices on UCMerced dataset}
	\label{fig:ConfusionMatrix4UCMerced}
\end{figure*}

In Table \ref{table1}, we can see that our proposed LEDL algorithm is only similar with CRC and still outperforms the other methods. Compared with LC-KSVD, LEDL achieves the higher accuracy by an improvement of $1.3$$\%$.
Here, we set $\lambda = {2^{ 0}}$, $\omega = {2^{ - 9}}$, $\varepsilon = {2^{ - 6}}$ to get the optimal result. The confusion matrices of the UC Merced Land Use dataset for all classes are shown in Figure \ref{fig:ConfusionMatrix4UCMerced}. We can see that, in all classes except the tennis, LEDL almost achieve better results compared with LC-KSVD. In several classes such as building, freeway, river, and sparse, our method achieves superior performance to LC-KSVD by an improvement of at least $0.5$$\%$.

\subsection{AID Dataset}
\begin{figure}[h!]
	\begin{center}
		\includegraphics[width = 0.8\linewidth]{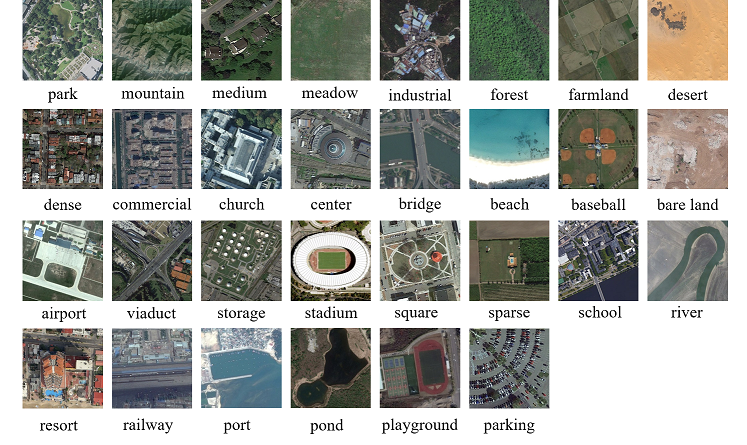}
	\end{center}
	\caption{Examples of the AID dataset}\label{fig:AID}
\end{figure}

The AID dataset is a new large-scale aerial image dataset which can be downloaded from Google Earth imagery. It contains $10{,}000$ images from 30 aerial scene types. In Figure \ref{fig:AID}, we show several images of this dataset. 

Table \ref{table1} illustrates the effectiveness of LEDL for classifying images. We adjust $\lambda = {2^{ -6}}$, $\omega = {2^{ - 14}}$, $\varepsilon = {2^{ - 12}}$ to achieve the highest accuracy by an improvement of at least $0.3$$\%$ in the five algorithms. While compared with LC-KSVD, LEDL achieves an improvement of $2.7$$\%$.

\subsection{Caltech101 Dataset}
\begin{figure}[h!]
	\begin{center}
		\includegraphics[width = 0.8\linewidth]{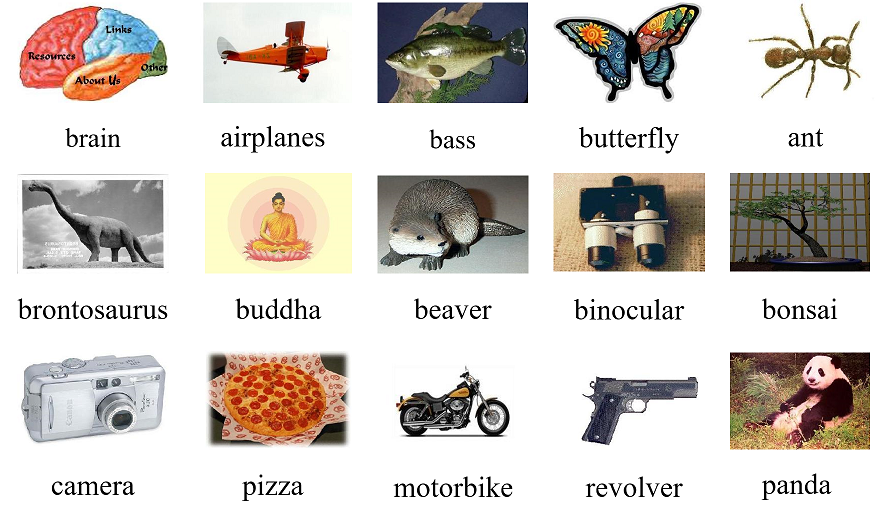}
	\end{center}
	\caption{Examples of the Caltech101 dataset}\label{fig:Caltech101}
\end{figure}

The caltech101 dataset includes $9{,}144$ images of $102$ classes in total, which are consisted of cars, faces, flowers and so on. Each category have about 40 to 800 images and most of them have about 50 images. In figure~\ref{fig:Caltech101}, we show several images of this dataset.

\begin{figure}[h!]
	\begin{center}
	\includegraphics[width=1.0\linewidth]{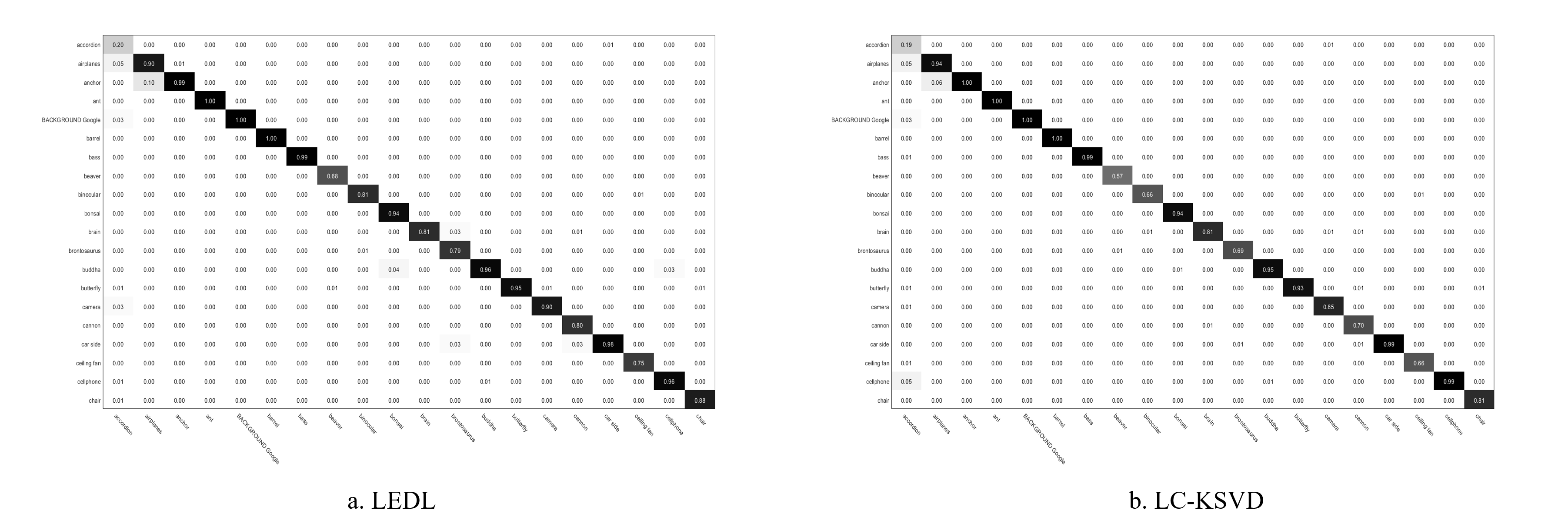}
\end{center}
\caption{Confusion matrices on Caltech101 dataset}
\label{fig:ConfusionMatrix4Caltech101}
\end{figure}

As can be seen in Table \ref{table1}, our proposed LEDL algorithm outperforms all the competing approaches by setting $\lambda = {2^{ - 4}}$, $\omega = {2^{ - 13}}$, $\varepsilon = {2^{ - 14}}$ and achieves improvements of $1.8$$\%$ and $0.7$$\%$ over LC-KSVD and other methods, respectively. Here, we also choose the first 20 classes to build the confusion matrices. They are shown in Figure~\ref{fig:ConfusionMatrix4Caltech101}.

\subsection{USPS Dataset}
\begin{figure}[h!]
	\begin{center}
		\includegraphics[width = 0.8\linewidth]{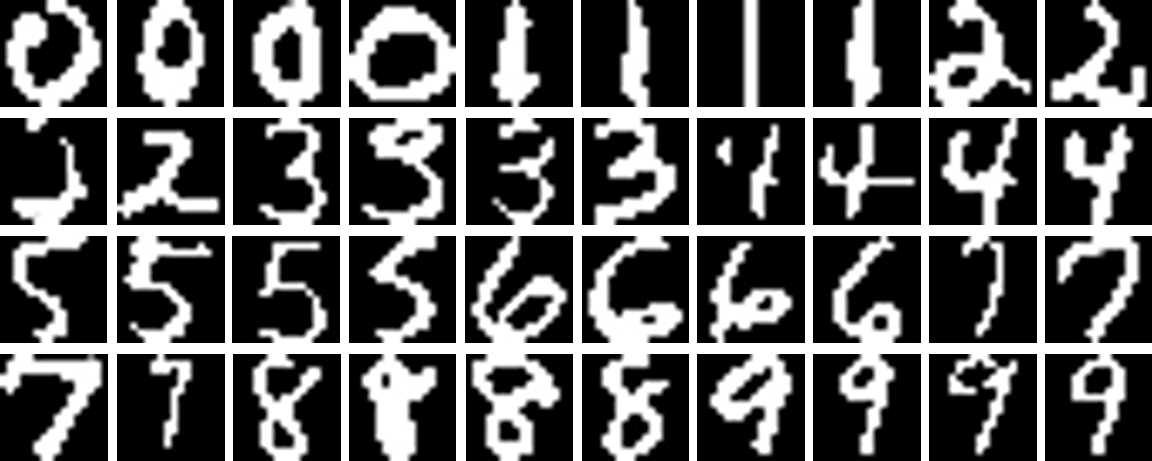}
	\end{center}
	\caption{Examples of the USPS dataset}\label{fig:USPS}
\end{figure}

The USPS dataset contains $9{,}298$ handwritten digit images from 0 to 9 which come from the U.S. Postal System. We list several samples from this dataset in Figure~\ref{fig:USPS}.

Table \ref{table1} shows the comparasion results of five algorithms and it is easy to find out that our proposed LEDL algorithm outperforms over other well-known methods by an improvement of at least $2.3$$\%$. And our proposed method achieves an improvement of $10.0$$\%$ over LC-KSVD method. The optimal parameters are $\lambda = {2^{ -4}}$, $\omega = {2^{ - 8}}$, $\varepsilon = {2^{ - 5}}$.

\subsection{Discussion}
From the experimental results on six datasets, we can obtain the following conclusions.

(1) All the above experimental results illustrate that, our proposed LEDL algorithm is an effective and general classifier which can achieve superior performacne to state-of-the-art methods on various datasets, especially on Extended YaleB dataset, CMU PIE dataset and USPS dataset.  

(2) Our proposed LEDL method introduces the $\ell_1$-norm regularization term to replace the $\ell_0$-norm regularization of LC-KSVD. However, compared with LC-KSVD algorithm, LEDL method is always better than it on the six datasets. Moreover, on the two face datasets and USPS dataset, our method can exceed LC-KSVD nearly $10.0$$\%$.

(3) Confusion matrices of LEDL and LC-KSVD on three datasets are shown in Figure~\ref{fig:ConfusionMatrix4ExtendedYaleB}~\ref{fig:ConfusionMatrix4UCMerced} and~\ref{fig:ConfusionMatrix4Caltech101}. They clearly illustrate the superiority of our method. Specificially, for Extended YaleB dataset, our method achieve outstanding performance in five classes (class1, class2, class3, class10, class16). For UC Merced dataset, LEDL almost achieve better classification rates than LC-KSVD in all classes except the tennis class. For Caltech101 dataset, our proposed LEDL method perform much better than LC-KSVD method in some classes such as beaver, binocular, brontosaurus, cannon and ceiling fan.

\section{Conclusion}
\label{Conclusion}

In this paper, we propose a Label Embedded Dictionary Learning (LEDL) algorithm. 
Specifically, we introduce the $\ell_1$-norm regularization term to replace the $\ell_0$-norm regularization term of LC-KSVD which can help to avoid the NP-hard problem and find optimal solution easily. 
Furthermore, we propose to adopt ADMM algorithm to solve $\ell_1$-norm optimization problem and BCD algorithm to update the dictionary. 
Besides, extensive experiments on six well-known benchmark datasets have proved the superiority of our proposed LEDL algorithm.

% if have a single appendix:
%\appendix[Proof of the Zonklar Equations]
% or
%\appendix  % for no appendix heading
% do not use \section anymore after \appendix, only \section*
% is possibly needed

% use appendices with more than one appendix
% then use \section to start each appendix
% you must declare a \section before using any
% \subsection or using \label (\appendices by itself
% starts a section numbered zero.)
%
\section{Acknowledgment}

This research was funded by the National Natural Science Foundation of China (Grant No. 61402535, No. 61671480), the Natural Science Foundation for Youths of Shandong Province, China (Grant No. ZR2014FQ001), the Natural Science Foundation of Shandong Province, China(Grant No. ZR2018MF017), Qingdao Science and Technology Project (No. 17-1-1-8-jch), the Fundamental Research Funds for the Central Universities, China University of Petroleum (East China) (Grant No. 16CX02060A, 17CX02027A), and the Innovation Project for Graduate Students of China University of Petroleum(East China) (No. YCX2018063).

% trigger a \newpage just before the given reference
% number - used to balance the columns on the last page
% adjust value as needed - may need to be readjusted if
% the document is modified later
%\IEEEtriggeratref{8}
% The "triggered" command can be changed if desired:
%\IEEEtriggercmd{\enlargethispage{-5in}}

% references section

% can use a bibliography generated by BibTeX as a .bbl file
% BibTeX documentation can be easily obtained at:
% http://mirror.ctan.org/biblio/bibtex/contrib/doc/
% The IEEEtran BibTeX style support page is at:
% http://www.michaelshell.org/tex/ieeetran/bibtex/
\bibliographystyle{elsarticle-harv}
\bibliography{egbib}
%
% <OR> manually copy in the resultant .bbl file
% set second argument of \begin to the number of references
% (used to reserve space for the reference number labels box)

% biography section
% 
% If you have an EPS/PDF photo (graphicx package needed) extra braces are
% needed around the contents of the optional argument to biography to prevent
% the LaTeX parser from getting confused when it sees the complicated
% \includegraphics command within an optional argument. (You could create
% your own custom macro containing the \includegraphics command to make things
% simpler here.)
%\begin{IEEEbiography}[{\includegraphics[width=1in,height=1.25in,clip,keepaspectratio]{mshell}}]{Michael Shell}
% or if you just want to reserve a space for a photo:

% if you will not have a photo at all:

% insert where needed to balance the two columns on the last page with
% biographies
%\newpage

% You can push biographies down or up by placing
% a \vfill before or after them. The appropriate
% use of \vfill depends on what kind of text is
% on the last page and whether or not the columns
% are being equalized.

%\vfill

% Can be used to pull up biographies so that the bottom of the last one
% is flush with the other column.
%\enlargethispage{-5in}

% that's all folks
\end{document}